# Automated Failure-Mode Clustering and Labeling for Informed Car-To-Driver Handover in Autonomous Vehicles

Aaquib Tabrez*
mohd.tabrez@colorado.edu
University of Colorado Boulder
Boulder, Colorado, USA

Matthew B. Luebbers*
matthew.luebbers@colorado.edu
University of Colorado Boulder
Boulder, Colorado, USA

Bradley Hayes
bradley.hayes@colorado.edu
University of Colorado Boulder
Boulder, Colorado, USA

## ABSTRACT
The car-to-driver handover is a critically important component of safe autonomous vehicle operation when the vehicle is unable to safely proceed on its own. Current implementations of this handover in automobiles take the form of a generic alarm indicating an imminent transfer of control back to the human driver. However, certain levels of vehicle autonomy may allow the driver to engage in other, non-driving related tasks prior to a handover, leading to substantial difficulty in quickly regaining situational awareness. This delay in re-orientation could potentially lead to life-threatening failures unless mitigating steps are taken. Explainable AI has been shown to improve fluency and teamwork in human-robot collaboration scenarios. Therefore, we hypothesize that by utilizing autonomous explanation, these car-to-driver handovers can be performed more safely and reliably. The rationale is, by providing the driver with additional situational knowledge, they will more rapidly focus on the relevant parts of the driving environment. Towards this end, we propose an algorithmic failure-mode identification and explanation approach to enable informed handovers from vehicle to driver. Furthermore, we propose a set of human-subjects driving-simulator studies to determine the appropriate form of explanation during handovers, as well as validate our framework.

## 1 MOTIVATION & BACKGROUND

The Society of Automotive Engineers (SAE) defines a framework through which autonomous driving features can be placed into one of six categories, ranging from Level 0 (full human control) to Level 5 (full autonomous control) [1]. Many autonomous vehicles (AVs) on the road today contain features approaching Level 3 in this taxonomy, commonly referred to as *eyes off* driving. In SAE Level 3, the vehicle assumes responsibility for all aspects of a certain driving modality (say, highway driving), allowing the human driver to focus on other tasks, but with the expectation that the driver must respond to a request to intervene at any time.

---

*These authors contributed equally to this work.

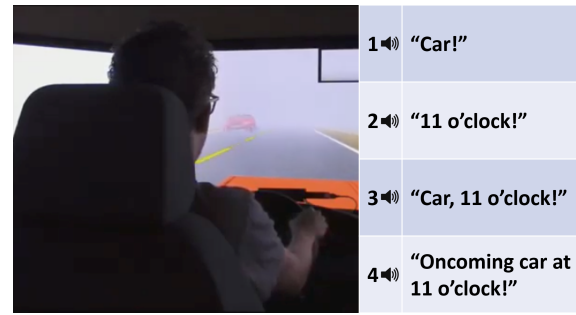

**Figure 1:** Proposed human-subjects study to evaluate different modalities for informing a driver of a potential failure state prior to handover in an autonomous vehicle simulator. Driving image from van der Heiden et al. [15]

Such car-to-driver control handovers occur whenever the autonomous driving system reaches its bounds of operational safety. However, this transfer of driving responsibility is by no means trivial; as drivers are permitted to be more distracted from the driving task, it takes them longer to regain the situational awareness needed to safely assume control [4, 5], and avoid the potentially dangerous state that caused the self-driving system to lose confidence, especially when that state arises from extrinsic factors (e.g., weather conditions, road elements, or traffic situations).

Prior work on the handover problem has investigated the benefits of auditory *pre-alerts*: series of beeps that precede the transfer to manual control by several seconds. Results from this study showed that pre-alerts cause the driver to disengage earlier from their assigned non-driving related task and focus more on the road prior to handover in comparison with the industry-standard brief alert followed by immediate transfer condition [15]. Although pre-alerting minimizes surprise, the driver must still discover the reason for handover on their own. Additionally, the approach does not provide any algorithmic apparatus for predicting the handover in advance, thus triggering the pre-alerts. Another study examining the handover problem by Du et al. revealed the effects of emotions on driver takeover, showing that positive emotional valence leads to improved post-takeover performance [3].

Preliminary research has been conducted into the use of explanations to engender trust in AVs as the vehicle is forced to perform unexpected actions or cede control [9, 16]. Explainable AI has emerged as a necessary component of fieldably safe autonomous systems in safety-critical domains, especially for the establishment of shared mental models and appropriate trust [2, 6, 8, 12, 13]. Therefore, we hypothesize in this work that utilizing explainable AI will lead to better informed and more fluent car-to-driver handover. To test this hypothesis, we propose, in the following sections, both an algorithmic approach aimed at informed handover and a series of user studies to determine its effectiveness.

## 2 APPROACH

Here, we detail a generalized theoretical framework of failure mode clustering and labeling for informed autonomous system-to-operator handover arising from uncertainty in that system's model and decision making.

The central insight behind making this handover *informed* is automated identification and explanation of anticipated risks using natural language to minimize surprise while maximizing handover fluency. The framework can be characterized by two components responsible for: i) automated clustering of anticipated risks, and ii) human intelligible and rapidly understandable labeling of predicted failures as explanations during handover.

*Automated Failure-Mode Clustering:* Our method of failure-mode clustering consists of a three-step procedure: 1) running Monte Carlo (MC) simulation rollouts within the configuration space to obtain expected rewards, 2) clustering rewards using Expectation-Maximization (EM), and 3) calculating the expected probabilities of failure-modes using occupational frequency (a form of histogram) across the MC rollouts.

First, given a model of the autonomous system and its environment, we randomly sample $k$ control inputs from this distribution and simulate $k$ forward rollouts of the system ($k$ in this case is a hyper-parameter informed by the complexity of the domain, accuracy of the desired output, and time required for the operator to gain situational awareness). For each pass, the solution is obtained in the form of a tuple containing the state and expected reward.

Then, we run the expectation-maximization algorithm to cluster the expected rewards from each MC rollout and obtain a set of states with similar expected rewards, modeled as Gaussian mixture models (GMMs).

Finally, with GMMs representing each potential reward cluster, we can use the final results from MC sampling to obtain the occupational frequency over the clusters. This probability distribution will be used to calculate the likelihood of each cluster, and via a thresholding mechanism, sufficiently likely clusters will have their mean reward computed, with low values indicating a potential failure-mode, triggering the labeling process.

*Human Intelligible Labeling of Failure-Mode:* At the end of the clustering phase, we have a set of clusters with associated likelihoods and mean rewards. Using these values, we trigger the labeling process via a pair of thresholding criteria. We approach the problem of labeling clusters as a set cover problem, trying to find the smallest logical expression of predicates that succinctly describe a cluster of states similar to Hayes and Shah [7]. Prior work used the Quine-McCluskey algorithm (QM) to resolve an arbitrary set of states into a natural language description by using a Boolean algebra over the space of defined predicates. Though this approach is innovative in performing explanation generation, the solution is exponential in memory and runtime complexity relative to the size of the domain and predicate set, preventing its use in most real-world problems. Unlike the prior work, we propose solving the problem of minimum set cover using an Integer Programming formulation [14], which shows multiple order of magnitude improvements over the state-of-the-art [7].

## 3 EXPERIMENTAL VALIDATION

To test our central hypothesis, we propose a human-subjects study (*Study 1*) using a driving simulator to compare handover alert modalities, and to validate our approach, we propose another driving simulator study (*Study 2*) to compare the performance of our algorithmic framework against the current state of the art.

***Study 1*** involves a simulated autonomous driving scenario, in which the human driver is provided with a distractor task to take their attention off of the road. Prior to a preformulated car-to-driver handover event, the vehicle will audibly alert the driver either with **a)** a generic alarm or phrase (e.g. "Handing over control!"), or **b)** a phrase with a semantic description of the failure state. Condition **b** is further broken down by modality of expression; **b.1)** identifies the object causing the failure state, **b.2)** identifies the direction relative to the driver in which that object lies, **b.3)** combines the information conveyed in b.1 and b.2, and **b.4)** presents the information of b.3 in the form of a complete sentence. For a representative example of the described conditions, see Figure 1.

Based on both the subjective measure of perceived helpfulness, and the objective measure of crash avoidance rate in the simulation, we hypothesize as follows: H1.) condition **b** will generally outperform **a**, H2.) condition **b.3** will outperform conditions **b.1** and **b.2**, because of the increase in conveyed information, and H3.) condition **b.3** will also outperform condition **b.4**, as the relative increase in information is not sufficient to overcome the increase in time overhead. Previous studies in cognitive science and psychology have shown that humans are adept at extracting semantic information and retaining it short-term even with non-complete sentence structure, giving preliminary insight for H3 [10, 11].

***Study 2*** would involve taking the best performing modality from *Study 1*, and using it to inform the implementation of our algorithmic formulation (Section 2). We would take the same simulated autonomous driving handover scenario as described in *Study 1*, and compare the effectiveness of a handover alarm triggered by our algorithm against the current state of the art (no information conveyed beyond the alarm itself). We hypothesize that the addition of information associated with our algorithmic approach will lead to objectively improved performance in avoiding failure states, as well as improved user perception of system safety and helpfulness.

## 4 CONCLUSION

In this paper, we have argued that explainable AI can be leveraged to improve car-to-driver handover in autonomous vehicles. In support of this, we have presented an algorithmic formulation enabling failure-mode identification and explanation for autonomous system-to-operator handover, capable of providing value in human-robot teaming domains, enabling mental model convergence for effective task fluency. We also proposed a set of human-subjects studies to inform the algorithmic implementation and later compare it to the state of the art.